\documentclass[11pt]{article}

\usepackage{acl}

\usepackage{times}
\usepackage{latexsym}

\usepackage[T1]{fontenc}

\usepackage[utf8]{inputenc}

\PassOptionsToPackage{x11names}{xcolor}
\usepackage[x11names]{xcolor} 
\usepackage{microtype}
\usepackage{todonotes}
\usepackage{graphicx}
\usepackage{booktabs}
\usepackage{soul}
\usepackage{enumitem}
\usepackage{expex}
\usepackage{makecell}

\usepackage{annotates}
\let\annotatescomment\comment %
\let\comment\undefined   
\usepackage[most]{tcolorbox}  %
\let\comment\annotatescomment %

\newtcbox{\greenbox}[1][]{nobeforeafter,tcbox raise base,colframe=white!100!black,colback=green!20!white, sharp corners,top=0pt,bottom=0pt,before upper=\strut,#1,right=0mm,left=0mm, boxrule=0mm}
\newtcbox{\bluebox}[1][]{nobeforeafter,tcbox raise base,colframe=white!100!black,colback=blue!20!white, sharp corners,top=0pt,bottom=0pt,before upper=\strut,#1,right=0mm,left=0mm,boxrule=0mm}

\title{\textit{EuskañolDS}: A Naturally Sourced Corpus for Basque-Spanish Code-Switching}

\author{Maite Heredia \quad Jeremy Barnes \quad Aitor Soroa \\
  HiTZ Center - Ixa, University of the Basque Country UPV/EHU \\
\texttt{\{maite.heredia\}@ehu.eus} \\
}

\begin{document}
\maketitle
\begin{abstract}
\textit{Code-switching} (CS) remains a significant challenge in Natural Language Processing (NLP), mainly due a lack of relevant data. In the context of the contact between the Basque and Spanish languages in the north of the Iberian Peninsula, CS frequently occurs in both formal and informal spontaneous interactions. However, resources to analyse this phenomenon and support the development and evaluation of models capable of understanding and generating code-switched language for this language pair are almost non-existent. We introduce a first approach to develop a naturally sourced corpus for Basque-Spanish code-switching. Our methodology consists of identifying CS texts from previously available corpora using language identification models, which are then manually validated to obtain a reliable subset of CS instances. We present the properties of our corpus and make it available under the name \textit{EuskañolDS}.\footnote{\textit{Euskañol} is a colloquial portmanteau used to describe the mixing of Basque and Spanish, whose endonyms are \textit{euskara} and \textit{español}, respectively.}

\end{abstract}

\section{Introduction}

Code-switching (CS) is a phenomenon that consists of alternating or mixing between two or more languages within a single discourse. It is common in multilingual communities, both in oral and written communication \citep{language-contact-appel-muysken,interlingual-online}. Since it is estimated that multilingualism is a standard for more than half of the world population \citep{global-bilingualism-tucker}, code-switching can be of great interest, alongside other phenomena that arise from language contact, such as borrowings and calques. Nevertheless, language models generally perform worse in CS scenarios, even multilingual models that are supposedly proficient in the languages \citep{winata-etal-2021-multilingual,aguilar-etal-2020-lince}. Therefore, it is essential to research and develop resources and applications for Natural Language Processing (NLP) that take into account the existence of CS.

According to the survey by \citet{winata-etal-2023-decades}, works on computational approaches to CS include Hindi-English, Spanish-English and Chinese-English as the most researched language pairs, although they point out the need to ``\textit{broaden the language scope of CS research}". In this respect, our research centres on Basque (ISO 639-3: eus), a minoritized low-resource language spoken in the in the western Pyrenees. Most of its speakers are bilingual and also speak Spanish (ISO 639-3: spa) or French (ISO 639-3: fra). The contact between these languages results in different common phenomena, including code-switching with either language, shown in examples \ref{ex:cs_es-eu} and \ref{ex:cs_fr-eu}. Although there are numerous studies on socio-pragmatic features of this contact \citep{Barredo2003PRAGMATICFO,vasco-castellano,ergative2020irantzu} and the presence of code-switching between Basque and Spanish is well documented in NLP \citep{garcia-sardina-etal-2018-es-port,escribano-etal-2022-basqueparl},  naturally sourced resources are still lacking.

\begin{table*}[h]
    \centering
    \resizebox{\linewidth}{!}{
    \begin{tabular}{lrll}\hline
       \textbf{Name} &\textbf{Size(Tokens)}& \textbf{Source} & \textbf{Topics}\\ \cmidrule(lr){1-1}\cmidrule(lr){2-2}\cmidrule(lr){3-3}\cmidrule(lr){4-4}
        BasqueParl & 14M & Parliamentary transcriptions & Political discourse \\
        HelduGazte & 37M & Twitter & \makecell[l]{News, sport, music, nationalist left}  \\
        Covid-19 &  57M & Twitter (September 2019 to February 2021) & Covid-19, political issues \\\hline
    \end{tabular}
    }
    \caption{Summary of the sources of the dataset. Topics correspond to those reported in their corresponding papers.}
    \label{tab:initial-sets}
\end{table*}

 \ex \label{ex:cs_es-eu}
 \textit{Y si lo ponen más bonito} eta musika polita jende gehiago juteko[...]!\\
    \textit{And if they decorate the place} and play good music, is so that people will go there[...]!\\
 \null\hfill\citep{Barredo2003PRAGMATICFO}
 \xe
 
  \ex \label{ex:cs_fr-eu}
 \textit{Comment sans boire de vin} egon behar dugu, \textit{nos malheureuses gorges} behar dute idortu.\\
    \textit{Since} we have to live \textit{without drinking wine, our unfortunate throats} will get dry.\\
 \null\hfill\citep{ergative2020irantzu}
 \xe

 To better study the contact between these languages and the ability of models to process CS, we develop a semi-supervised methodology to obtain code-switched sentences from pre-existing corpora, with the help of language identification models. We use our approach to gather a corpus from diverse sources--parliamentary transcriptions and social media posts--and present some qualitative analysis on its contents.  We make our dataset and the code used to gather and process it publicly available.\footnote{URL to be announced upon acceptance.} 

\section{The EuskañolDS Dataset}
We present a dataset of instances containing code-switching between Spanish and Basque, which are two languages from different linguistic families with many perceptible typological differences. For example, Spanish is an fusional language with nominative-accusative alignment and SVO dominant word order, whereas Basque is an agglutinative language with ergative-absolutive alignment, and its dominant word order is SOV. Their similarities include similar phonetic inventories, and a long shared history of contact and mutual sociolinguistic influence. 

The final \textit{EuskañolDS} dataset has two splits: silver, automatically classified, and gold, manually filtered. We explain in detail the process to obtain them, as well as provide a quantitative and qualitative analysis on the interesting properties of our dataset.

\subsection{Dataset Sources}
We source our data from the following corpora, summarized in Table \ref{tab:initial-sets}: 
\paragraph{BasqueParl \citep{escribano-etal-2022-basqueparl}} is a corpus of Basque parliamentary transcriptions. Basque and Spanish are both vehicular languages in these interactions, resulting in "\textit{heavy Basque-Spanish code-switching}", as well as frequent language switches between exchanges. 
\paragraph{HelduGazte \citep{FernandezdeLanda2019LargeSL,heldugazte}} is composed of almost 6 million tweets by Basque speakers, used to analyse the use of formal and informal Basque on social media, as well as adult and young speech.\footnote{We have only been able to access 1000 tagged tweets from this corpus, because the rest of the instances are not available without access to the X API.} 
\paragraph{Covid-19 \citep{fernandez-de-landa-etal-2024-uncovering}} is a corpus of 8 million tweets by Basque speakers during the Covid-19 pandemic, used to study diachronic trends of language use during different stages of the pandemic.

\paragraph{}We consider BasqueParl as a source of formal language, whereas both twitter datasets include a mixture of formal and informal speech, tending towards informal. Therefore, our corpus collects a wide variety of topics and different diaphasic and diatopic varieties, as well as spoken and written language.

\subsection{Silver Set: Automatic Classification}
To filter the instances, we propose a semi-supervised approach that first employs the \href{https://huggingface.co/facebook/fasttext-language-identification}{model for Language Identification} from FastText \citep{joulin2016fasttext,joulin2016bag} to automatically classify the instances. The model was trained to identify 217 languages, including Spanish and Basque, and also outputs a confidence level for each tag. In our dataset, both the average and the median confidence of the predictions are 99\%, indicating that the majority of the predictions of the models have a high confidence. When instances are filtered by their confidence level, the lower the confidence, the higher probability of them containing CS. Preliminary testing indicated that filtering instances that have a confidence lower than 90\% and that are tagged as Basque and Spanish or viceversa gives us a high-precision set of instances exhibiting CS.

The final \textbf{silver split} has a total of $20,008$ instances, $597$ sourced from BasqueParl, $19,339$ from Covid-19 and $72$ from HelduGazte.

\subsection{Gold Set: Manual Validation}
To obtain a gold-standard test set, we manually verify a subset of the automatically filtered sentences. We classify all BasqueParl and HelduGazte instances and 2000 random instances from the Covid-19 corpus, to balance the texts from both sources. In order to distinguish CS from similar phenomena such as borrowings \citep{alvarez-mellado-lignos-2022-detecting}, we only consider sentences that contain more than two words in each language and grammatical features from both languages as CS, although this aggressive filtering removes utterances that could be considered as code-switching. We also do not consider CS instances where the switch occurs at a proper noun that has no direct translation, as in Example \ref{ex:negative-example1}, or where the content of both languages is the same, as in Example \ref{ex:negative-example2}. 

   \ex \label{ex:negative-example1}
    La candidata de EH Bildu es Maddalen Iriarte, documentate.\\
   Maddalen Iriarte is EH Bildu's candidate, get informed.
 \xe  
 \ex \label{ex:negative-example2}
    Dublin, gaur ! . Ederra benetan!! / \textit{Dublin, hoy. Es precioso!}\\
    Dublin, today ! . Truly beautiful!! / \textit{Dublin, today. It's beautiful!}
 \xe
 The final \textbf{gold split} has a total of $927$ manually filtered instances, $403$ sourced from BasqueParl, $72$ from HelduGazte, and $452$ from the Covid-19 corpus. 
\begin{table}[t]
    \centering
    \resizebox{\linewidth}{!}{
    \begin{tabular}{lrrr}\hline
       \textbf{Split} &\textbf{Tokens}&\textbf{Instances}&\textbf{Avg. Length}\\ 
       \cmidrule(lr){1-1}\cmidrule(lr){2-2}\cmidrule(lr){3-3}\cmidrule(lr){4-4}
        Silver & 537,648  & 20,008 & 26.87\\
        Gold & 36,860  & 927 & 39.76\\
        \hline
    \end{tabular}}
    \caption{Quantitative analysis of \textit{EuskañolDS}.}
    \label{tab:quantitative-analysis}
\end{table}
\begin{table*}[t!]
    \centering
    \resizebox{\linewidth}{!}{
    \begin{tabular}{cccc}
        \hline \textbf{Source}&\textbf{Instance}& \textbf{Translation}&\textbf{Type of CS}\\
        \cmidrule(lr){1-1}\cmidrule(lr){2-2}\cmidrule(lr){3-3}\cmidrule(lr){4-4}HelduGazte&\greenbox[rounded corners = west]{bihar zazpi terditan gora}\bluebox[rounded corners = east]{y yo me muerooooooo} & tomorrow up at seven thirty \textit{and i'm going to die}&Intra-sentential\\
       BasqueParl&\makecell[c]{\bluebox[rounded corners = west]{Por lo tanto, no tengo nada más que añadir. }\\\greenbox[rounded corners = east]{Eta eskerrik asko denoi akordio batera heldu garelako.}} &  \makecell[c]{\textit{Therefore, I don't have anything else to add.}\\And thank you everyone for having reached an agreement.}&Inter-sentential\\ Covid-19&\greenbox[rounded corners = west]{Katxis!}\bluebox{Veo a la tropa baja...}\greenbox[rounded corners = east]{Eutsi goiari!} & Heck! \textit{I see the spirits are low...} Cheer up!&Emblematic \\
        \hline
    \end{tabular}
    }
    \caption{Examples from the dataset. Basque in green, Spanish in blue.}
    \label{tab:examples}
\end{table*}

Table \ref{tab:quantitative-analysis} shows some quantitative statistics from our corpus, comparing the size of both splits. Although the silver set has 20 times more instances than the gold set, it has fewer tokens per instance on average, because it has a larger proportion of tweets, which are much shorter on average.

\subsection{Qualitative Analysis}
As a first insight into our corpus, we perform a manual analysis of some qualitative aspects of our gold set. First, we classify the instances according to the following widespread typology \citep{Appel_Muysken_2006}, illustrated in Table \ref{tab:examples} with instances from the corpus:

\begin{itemize}
    \item \textbf{Inter-sentential CS} occurs between sentences, and is the most represented type in our corpus.
    \item \textbf{Intra-sentential CS} occurs in the middle of a sentence. %
    \item \textbf{Emblematic CS} occurs between a sentence and an exclamation or a tag.
\end{itemize}

As shown in Table \ref{tab:cs-types}, most instances exhibit inter-sentential CS, mainly due to those from Covid-19 and BasqueParl. On the other hand, HelduGazte has more intra-sentential or code-mixed sentences, but also comes from a smaller and less representative dataset. The least represented type is emblematic CS, 3.14\% of the total corpus. This proportion may have been larger if we included more informal conversations, as they tend to occur more often in informal oral speech \citep{Ibarra_Murillo_2019}.

\begin{table}
    \centering
    \begin{tabular}{crrr}
        \cmidrule(lr){1-4}
         & \textbf{Inter} & \textbf{Intra} & \textbf{Emblem} \\
         \cmidrule(lr){2-2}\cmidrule(lr){3-3}\cmidrule(lr){4-4}
         HelduGazte & 36.11\% & 58.33\% & 5.56\%\\
         Covid-19 & 85.40\% & 9.73\% & 4.87\%\\
         BasqueParl & 67.25\% & 31.76\% & 0.99\%\\\midrule
         Total & 73.68\% & 23.09\% & 3.24\% \\\bottomrule
    \end{tabular}
    \caption{Proportion of each type of CS in the gold split according to the source of the instances and in total.}
    \label{tab:cs-types}
\end{table}

The tweets in our dataset often contain both informality traits and dialectal elements. The presence of different Basque dialects, also called \textit{euskalkiak}, is specially notable. In Example \ref{ex:euskalki}, we can see some of these traits: compare standard \textit{temporada} with \textit{temporadie} or \textit{dago} with \textit{dao}.

  \ex \label{ex:euskalki}
    Ezteu nahi bezela hasi tenporadie, baño hau hasi besteik ezta eñ ta lan asko daola etteko garbi dao. \textit{Un placer volver a ver tantas caras conocidas.}\\
    I don't want to start the season like this, but not only has it just started and it is clear that there is a lot to do. \textit{A pleasure to see so many familiar faces.}.
 \xe

Similarly, inter-sentential CS is common in reported speech, where the language shifts when reporting what someone said in a different language. This is especially true in BasqueParl, as the speakers are constantly referencing other interventions. 

  \ex \label{ex:fatica}
    Edo beste erantzun berean esandako beste gauza bat: ``\textit{Así el modelo A, en su distribución horaria actual tendría que reformularse}".\\
    And another thing said in that same answer: ``\textit{Thus, model A, in its current time distribution, would have to be reformulated.}".
 \xe

The nature of isolated tweets means that we are missing important context (responses, retweets, etc.) and metadata about the authors that could provide insights into trends and motivation behind code-switching. However, in some cases we can infer the speaker's intent based solely on the textual content. For example, following \citet{Appel_Muysken_2006}, who identify Jakobson's six functions of language with six possible motivations behind code-switching, we can see the phatic function in Example \ref{ex:fatica}, where code-switching is used to test the language of the interaction, or the expressive function in Example \ref{ex:expresiva}, where code-switching is used to emphasize the feeling expressed. 

  \ex \label{ex:fatica}
 Kaixo Aitor. Euskaraz bai? \textit{Hablas euskara? Es para ver si podemos hacerte una entrevista} [...].\\
    Hi Aitor. Is Basque okay? \textit{Do you speak Basque? We would like to know if we can interview you}  [...].
 \xe
 
   \ex \label{ex:expresiva}
 \textit{Sencillamente alucinante.} Izugarria. Komentariorik ez...\\
    \textit{Just awesome.} Incredible. No comments...
 \xe

\section{Applications \& Future Work}
The dataset we have presented is the first resource that gathers instances with Basque-Spanish CS, and represents a first step towards evaluating and training models for this language pair. 

This corpus could be used for the theoretical study of code-switching features between Basque and Spanish, as demonstrated by the shallow insights already provided here. It can also be useful to develop datasets for NLP tasks, such as token language identification or stance detection on CS text, either on its own or in combination with other monolingual or bilingual datasets. %

\section{Conclusion}
In this paper, we present \textit{EuskañolDS}, a new resource for Basque-Spanish code-switching. It consists of a corpus of 20,000 instances sourced from tweets and parliamentary transcriptions. The instances have been filtered with a Language Identification system and manually classified, resulting in two versions of the corpus: a silver set, that contains all of the automatically identified instances, and a gold set, that only contains a reliable subset. We also present a first exploration of the phenomena observed in the corpus. We believe it is a resource of interest for different NLP and linguistic applications, that can open the door for both practical and theoretical research in Basque-Spanish CS.

\section*{Limitations}
The limitations that we have encountered during the creation of our corpus are mainly related to the low-resource status of Basque and the limited previous research on the Basque-Spanish CS pair in the NLP field. 
The data collection was made available thanks to previous works of researchers to gather natural corpora for the Basque language. The corpora of Basque tweets is specially relevant, because the X API  has since been closed, limiting the availability of spontaneous data that includes not only instances of CS but also other research topics.
Finally, we would like to mention that our corpus only refers to code-switching between Basque-Spanish, as we have considered the Basque-French pair to be out of scope for the current work.

\bibliography{anthology}
\bibliographystyle{acl_natbib}

\appendix

\end{document}